# Causal Conclusions that Flip Repeatedly and Their Justification

**Kevin T. Kelly and Conor Mayo-Wilson**
Department of Philosophy
Carnegie Mellon University
kk3n@andrew.cmu.edu and conormw@andrew.cmu.edu

**Abstract**

Over the past two decades, several consistent procedures have been designed to infer causal conclusions from observational data. We prove that if the true causal network might be an arbitrary, linear Gaussian network or a discrete Bayes network, then *every* unambiguous causal conclusion produced by a consistent method from non-experimental data is subject to *reversal* as the sample size increases *any finite number of times*. That result, called the *causal flipping theorem*, extends prior results to the effect that causal discovery cannot be reliable on a given sample size. We argue that since repeated flipping of causal conclusions is unavoidable in principle for consistent methods, the best possible discovery methods are consistent methods that retract their earlier conclusions no more than necessary. A series of simulations of various methods across a wide range of sample sizes illustrates concretely both the theorem and the principle of comparing methods in terms of retractions.

## 1 Introduction

Over the past two decades, several procedures have been designed to infer causal conclusions from observational data. However, there is an essential difference between causal conclusions based on experiments and those inferred from non-experimental data. Consider a randomized experimental trial to determine whether $X$ causes $Y$. Following the usual logic of statistical testing, one can suspend judgment regarding the relationship between $X$ and $Y$ until a statistically significant correlation is detected. If sample size is not sufficiently large, one can safely conclude that the missed effect is too small to be of practical consequence even if it exists. The same cannot be said of causal relations discovered, even unambiguously, from non-experimental data when the underlying causal truth is assumed to be linear gaussian or a discrete Bayes net (Spirtes et al. 2000, p. 83). Because the orientation of arbitrarily large causes can be retracted or reversed as sample size increases in these cases, it is impossible to find non-trivial interval estimates of the effect of a policy; depending on the orientation of the relevant causal connection, the result of the policy might either be null or quite large (Robins et al. 1999).

In this paper, we extend the result in (Robins et al. 1999) that causal discovery cannot be reliable on a given sample size. We prove that every causal conclusion drawn by a consistent causal discovery procedure can flip in orientation *any number of times* and with *arbitrarily high chance* as the sample size increases, as long as the truth might be a linear Gaussian network or a discrete Bayes network. That is the case regardless of the sample size at which the first conclusion in the sequence is produced. It is also true regardless of the strength of the causal connection between $X$ and $Y$, so the practical consequences of the flips could be immense. Finally, it remains true even if the truth *might* be an arbitrary, linear Gaussian or a discrete Bayes net. One application of the causal flipping theorem is that the two causal flips of the arrow between $X$ and $Y$ as depicted in figure 1 are unavoidable, regardless of the parameter settings of the first network in the sequence, if the causal discovery procedure is consistent and if the networks are either linear Gaussian or discrete. We illustrate the causal flipping theorem by showing in simulations how four published causal discovery algorithms perform the causal flips in figure 1 with high chance when provided with random samples of increasing size generated from a *fixed* setting of the free parameters in the third causal network in the series.

We do not infer from the causal flipping theorem that causal discovery is unjustified and should, therefore, be abandoned. Rather, we view the unavoidability of causal flipping as the proper *justification* for causal discovery, in the sense that the best causal discovery algorithms are those that *minimize* flipping prior to convergence to the true model. It can be shown (Kelly 2010, Kelly and Mayo-Wilson 2010, Kelly 2007b) that causal inference methods that systematically prefer simpler models (where simplicity is measured in terms of the number of conditional dependencies entailed by the model, and hence, by number of edges) are exactly those that also minimize retractions of earlier judgments. That result, called the **Ockham Efficiency Theorem**, provides a theoretical justification for existing algorithms that favor simpler models; a justification that does not depend in any way on question-begging, simplicity-biased prior probabilities. Moreover, methods that retract their conclusions less can be said to improve upon methods that retract more, so retraction minimization provides an objective standard of progress in the design of causal discovery algorithms. The simulations at the conclusion of the paper illustrate concrete comparisons for some published methods.

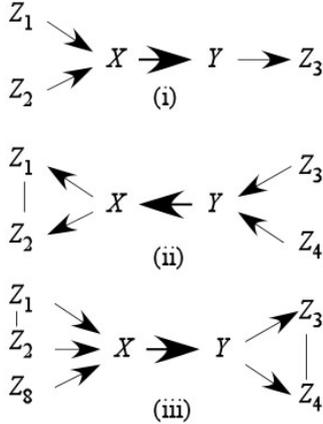

Figure 1: Causal Flips

## 2 Statistical Questions and Consistent Methods

Let $\mathbf{P}$ be a set of probability measures over a common $\sigma$-field. A **statistical question** with **presupposition $\mathbf{P}$** is a partition $\Theta$ of $\mathbf{P}$ into a countable collection of mutually exclusive and exhaustive **theories**. The aim is to find the theory $\mathbf{T}_p$ in $\Theta$ that contains the unknown, true probability measure $p$ by means of samples from $p$. For example, one might ask whether the mean of an unknown, one-dimensional normal distribution of standard variance is or is not exactly 0. In that case, let $\mathbf{P}$ be the set of all normal measures of standard variance, let $\mathbf{T}_{\mu=0}$ be the set of all such measures with mean 0, and let $\mathbf{T}_{\mu\neq 0}$ be the set of all measures in $\mathbf{P}$ with non-zero mean. Then the question corresponds to the partition $\Theta_\mu = \{\mathbf{T}_{\mu=0}, \mathbf{T}_{\mu\neq 0}\}$.

Define a **method** for $\Theta$ to be a mapping $M$ from i.i.d. samples to theories in $\Theta$. If $p \in \mathbf{P}$, let $p^n$ denote the sampling density over i.i.d. samples of size $n$. Typically, as in the question whether $\mu = 0$, there is no way to bound the chance of error at a given sample size over all of $\mathbf{P}$, but one can at least choose a method that converges to the truth in probability. Say that $M$ is **consistent** for $\Theta$ if and only if for each $p$ in $\mathbf{P}$ and for each $\epsilon > 0$, there exists sample size $n$ such that for each sample size $m \geq n$, it is the case that $p^m(M = \mathbf{T}_p) > 1 - \epsilon$. Consistency is a weak property, but it is the most that has been claimed for many published algorithms for causal discovery (Spirtes et al 2000). Our proposal, described in greater detail below, is that such discovery procedures are motivated and justified better if one *refines* the concept of consistency by minimizing fluctuations in the chances of producing alternative theories prior to convergence down to those that are unavoidable in principle.

## 3 Causal Discovery

It is a familiar fact that causal questions are not necessarily reducible to statistical dependency, since statistical dependency between $X$ and $Y$ is compatible with $X$ being a cause of $Y$, with $Y$ being a cause of $X$, or with a common cause of $X$ and $Y$. One of the bold insights of the recent literature on causal discovery is that, nonetheless, *some* interesting causal conclusions are determined entirely by probability, under some plausible assumptions. In this section, we present the most basic ideas behind the approach in order to fix notation—extended, motivated presentations may be found in (Pearl and Verma 1991, Spirtes et al. 2000). Let $\mathcal{V}$ be a fixed, finite set of random variables and let $\mathbf{P}_\mathcal{V}$ denote the set of all probability measures induced by joint probability densities over the variables in $\mathcal{V}$. Let DAG denote the set of all directed acyclic graphs on the fixed variable set $\mathcal{V}$. The arrows or **directed edges** in these DAGs are understood to indicate causal influence (as distinct from mere statistical dependence).

The key idea behind causal discovery is to assume some plausible principles that connect causal structure with the patterns of statistical dependencies. Let $X \amalg Y \mid \mathcal{S}$ abbreviate that $X$ is probabilistically independent of $Y$ conditional on the values of variables in $\mathcal{S}$, where $\mathcal{S} \subseteq \mathcal{V} \setminus \{X, Y\}$. Call such a statement a **conditional independence constraint** (CIC) over $\mathcal{V}$, and let CIC denote the set of all CICs over $\mathcal{V}$. If $p \in \mathbf{P}_\mathcal{V}$, define the CIC **pattern** $Ip$ to be the set of all CICs satisfied by $p$. If $\mathcal{G} \in$ DAG, say that $p$ is **Markov** for $\mathcal{G}$ if and only if for all variables $X \in \mathcal{V}$, it is the case that $X$ is independent of its non-descendants conditional on its parents. Let the CIC **pattern** of DAG $\mathcal{G}$ be defined as $I\mathcal{G} = \bigcap\{Ip : p \in \mathbf{P}_\mathcal{V}$ and $p$ is Markov for $\mathcal{G}\}$. When $I\mathcal{G} = Ip$, say that $p$ is **faithful** to $\mathcal{G}$. Faithfulness is the assumed connection between probability and causation. Define:

$$\begin{aligned} \text{FCIC} &= \{I\mathcal{G} : \mathcal{G} \in \text{DAG}\}; \\ \mathbf{F} &= \{p \in \mathbf{P}_\mathcal{V} : (\exists A \in \text{FCIC})\ Ip = A\}. \end{aligned}$$

That is, $\mathbf{F}$ is the set of measures faithful to some DAG over $\mathcal{V}$. Let $A \in$ FCIC. Define:

$$\begin{aligned} \mathbf{G}_A &= \{\mathcal{G} \in \text{DAG} : I\mathcal{G} = A\}; \\ \mathbf{F}_A &= \{p \in \mathbf{F} : Ip = A\}. \end{aligned}$$

Then $\Theta_0 = \{\mathbf{F}_A : A \in \text{FCIC}\}$ is a purely statistical question over $\mathbf{F}$, since $Ip$ is a function of $p$. Theory $\mathbf{F}_A$ uniquely determines $A$ and, hence, the set of DAGs $\mathbf{G}_A$, so $\Theta_0$ has potential causal significance. If all of the DAGs in $\mathbf{G}_A$ share a feature (e.g., the causal arrow $X \to Y$), then that feature is entailed by the statistical hypothesis $\mathbf{F}_A$, so the statistical question $\Theta_0$ may have strong, causal implications.

Therefore, causal discovery decomposes naturally into two parts, the purely *empirical* problem of determining $Ip$ by sampling from $p$ and the purely *deductive* problem of recovering $\mathbf{G}_{Ip}$ from $Ip$. Much of the recent progress in causal discovery has centered on the deductive phase of the problem, which proceeds as follows. The **causal skeleton** skel($\mathcal{G}$) is just the undirected graph with vertices $\mathcal{V}$ that has a undirected edge $X - Y$ between $X$ and $Y$ if and only if $\mathcal{G}$ has an edge between $X$ and $Y$. A **vee** in $\mathcal{G}$ is a configuration of form $X - Y - Z$ such that $X$ and $Z$ are not adjacent. An **unshielded collision** is a vee oriented as $X \to Y \leftarrow Z$.

**Proposition 1** (Verma and Pearl 1991). *$I\mathcal{G} = I\mathcal{G}'$ if and only if skel($\mathcal{G}$) = skel($\mathcal{G}'$) and $\mathcal{G}, \mathcal{G}'$ have the same unshielded collisions.*

Thus, the causal skeleton of the true $\mathcal{G}$ and all of the unshielded collisions in $\mathcal{G}$ are recoverable uniquely from the purely probabilistic information $I\mathcal{G}$. It may also be possible to derive further conclusions about $\mathcal{G}$. If the vee $X \to Y - Z$ results from orienting some unshielded collision, then it follows that $X \to Y - Z$ is not a collision, and the only remaining possibility is $Y \to Z$. Also, DAGs are acyclic, so if there is a directed path from $X$ to $Y$ and an edge $X - Y$, then it is safe to conclude that $X \to Y$. These two conditions can be iterated until no more orientations can be obtained. The resulting method is called the SGS algorithm, after the initials of its inventors (Spirtes et al. 2000). The same authors have produced a several more efficient and more sophisticated variants of the same idea, called the PC, CPC, and FCI algorithms that will employed in the simulation studies below.

Often, the pertinent causal question concerns some *local* feature of causal structure, such as whether $X$ causes $Y$ or $Y$ causes $X$ or neither, rather than some *global* causal structure, like the CIC pattern, which concerns all the variables in $\mathcal{V}$. Say that causal edge $X \to Y$ is **essential** in $\mathcal{G}$ if for all $\mathcal{G}'$ such that $I\mathcal{G} = I\mathcal{G}'$, the graph $\mathcal{G}'$ contains the directed edge $X \to Y$. Define:

$$\begin{aligned} \mathbf{T}_{X \to Y} &= \bigcup\{\mathbf{F}_\mathcal{G} : X \to Y \text{ is essential in } \mathcal{G}\}; \\ \mathbf{T}_{X \leftarrow Y} &= \bigcup\{\mathbf{F}_\mathcal{G} : X \leftarrow Y \text{ is essential in } \mathcal{G}\}; \\ \mathbf{T}_{X - Y} &= \bigcup\{\mathbf{F}_\mathcal{G} : X \to Y \text{ or } X \leftarrow Y \text{ is in} \\ &\qquad \mathcal{G} \text{ non-essentially}\}; \\ \mathbf{T}_{\neg(X-Y)} &= \bigcup\{\mathbf{F}_\mathcal{G} : X \text{ and } Y \text{ are non-adjacent in } \mathcal{G}\}. \end{aligned}$$

Define the causal question $\Theta_{X,Y}$ concerning $X, Y$ to be

$$\Theta_{X,Y} = \{\mathbf{T}_{X \to Y}, \mathbf{T}_{X \leftarrow Y}, \mathbf{T}_{X-Y}, \mathbf{T}_{\neg(X-Y)}\}.$$

## 4 Causal Flipping

The preceding section focused entirely on the purely deductive inference from $Ip$ to $\mathbf{G}_{Ip}$. But $Ip$ must be inferred from samples. One approach is to assume that a given CIC is true until it is rejected by a statistical test. If the significance levels of the tests are adjusted downward at a sufficiently slow rate, then this procedure converges in probability to $Ip$. Application of the above deductive rules computes a pattern that uniquely determines $\mathbf{G}_{Ip}$. The resulting method $M_0$ is consistent for $\Theta_0$ (Spirtes et al. 2000).

The consistency property does not imply a non-trivial bound on chance of error in the short run (Robbins et al. 1999), but the situation is worse than that. It is possible for a consistent method to produce the causal conclusion $X \to Y$ with arbitrarily high chance at a given sample size and then to produce the contrary conclusion $X \leftarrow Y$ with arbitrarily high chance as sample size increases, and so on, any number of times as sample size increases. Each such reversal in chance of the inferred causal orientation is a **causal flip**. Causal flips are not so surprising for the SGS algorithm, since that algorithm bases its conclusions entirely on the pattern of currently accepted and rejected null hypotheses, and the accepted null hypotheses may be rejected with high chance at higher sample sizes. The causal flipping theorem, however, asserts the *unavoidability* of arbitrarily many causal flips, if the causal discovery method is consistent and if arbitrary, linear Gaussian or discrete Bayes networks are theoretical possibilities. In the linear Gaussian and discrete Bayes net cases, it is as if every consistent method is essentially forced to base its causal conclusions on inferred patterns of conditional independence, and the inferred independence hypotheses may always be overturned as the sample size increases. That includes methods based on consistent scoring rules like BIC (see the GES simulation below) or even methods based on Bayesian posterior probabilities.

Recent work (Hoyer et al. 2009) establishes that in some questions that *exclude* linear Gaussian models and discrete Bayes nets as possibilities, the true causal structure *can* be identified without risking any retractions of earlier conclusions. However, that approach avoids retractions by producing no causal conclusions whatever all until the departure from the linear Gaussian case is noticeable in the data. Therefore, unless one foregoes all causal conclusions in linear Gaussian models, causal flipping remains an unavoidable feature of causal discovery from non-experimental data.

## 5 Empirical Approximation

To prove the causal flipping theorem, we introduce a few crucial definitions. Let $\mathbf{P}$ be a set of probability measures on a common $\sigma$-field $\mathcal{F}$, and define the **total variation distance** between two measures in $p, q \in \mathbf{P}$ as $\rho(p,q) = \sup_{E \in \mathcal{F}} |p(E) - q(E)|$. Total variation distance is a very natural measure of indistinguishability, since it bounds the difference in chances the two measures assign to an arbitrarily chosen "acceptance zone" for a test that distinguishes them. **Total variation space** is the topology induced by open $\rho$-balls. When $\rho$ is applied to sampling densities in statistical applications, we assume that one first converts the sampling densities to their induced, infinite product measures.

Given $\mathbf{T}, \mathbf{T}' \subseteq \mathbf{P}$, define the **empirical approximation order** with respect to $\mathbf{P}$:

$$\mathbf{T} \leq_{\mathbf{P}} \mathbf{T}' \Leftrightarrow \mathbf{T} \subseteq \text{cl}(\mathbf{T}')$$

where the topological closure operation is defined with respect to total variation space. When $\mathbf{P}$ is clear from context, we write drop the subscript $\mathbf{P}$ on $\leq$. The empirical approximation order is one of the main ideas in this paper—it embodies the philosophical problem of induction as it arises in statistical inference. Suppose that $\mathbf{T} \leq \mathbf{T}'$. Then no matter how much evidence you think you have for concluding that the true distribution is in $\mathbf{T}$, there is an essentially *indistinguishable* distribution in $\mathbf{T}'$. For example, recall the question $\Theta_\mu = \{\mathbf{T}_{\mu=0}, \mathbf{T}_{\mu\neq0}\}$, which asks whether the mean of a univariate normal distribution of standard variance is or is not identically 0. Let $p_\mu$ be the univariate normal with variance 1 and $\mu$. The unique element of $\mathbf{T}_{\mu=0}$, then, is $p_0$. If $\delta > 0$ is sufficiently small, then $\rho(p_0, p_\delta) < \epsilon$. Thus, $\mathbf{T}_{\mu=0} \leq \mathbf{T}_{\mu\neq0}$.

Because $\mathbf{T}_{\mu=0} \leq \mathbf{T}_{\mu\neq0}$, any consistent method for $\Theta_\mu$ can be forced to first conjecture $\mathbf{T}_{\mu=0}$, with arbitrarily high probability, and then later conjecture $\mathbf{T}_{\mu\neq0}$, again with arbitrarily high probability, at some larger sample size. Why? Suppose that $M$ is a consistent solution to $\Theta_\mu$, and let $\epsilon > 0$ be small. Then there exists sample size $n$ such that $p_0^n(M = \mathbf{T}_{\mu=0}) > 1 - \epsilon/2$. Since $\mathbf{T}_{\mu=0} \leq \mathbf{T}_{\mu\neq0}$, there is some $p_\delta \in \mathbf{T}_{\mu\neq0}$ such that $\rho(p_0, p_\delta) < p_0^n(M = \mathbf{T}_{\mu=0}) - (1 - \epsilon/2)$. By the definition of $\rho$, it follows that $q^n(M = \mathbf{T}_{\mu=0}) > 1 - \epsilon/2$, since $M = \mathbf{T}_{\mu=0}$ is an event in $\mathcal{F}$. But then the consistency of $M$ in $p_\delta$ guarantees that $p_\delta^m(M = \mathbf{T}_{\mu\neq0}) > 1 - \epsilon/2$, for some $m > n$. Thus, the chance that $M$ produces $\mathbf{T}_{\mu=0}$ plummets by a whopping $1 - \epsilon$ from sample size $n$ to sample size $m$. The change of heart on increasing information is not a mere matter of sampling noise; it represents a fundamental reversal in the *signal* the method sends to the user regarding theory $\mathbf{T}_{\mu=0}$ and it is a consequence of nothing more than the (desirable) consistency of $M$ and an unavoidable structural fact about the question addressed, namely, that $\mathbf{T}_{\mu=0} \leq \mathbf{T}_{\mu\neq0}$.

So much is implicit in standard, textbook discussions of statistical power, but the argument *iterates* as sample size increases without bound. Suppose that one is confronted with a question $\Theta$ in which there is an **empirical approximation chain** $\mathbf{T}_0 \leq \mathbf{T}_2 \leq \ldots \leq \mathbf{T}_k$, where $\mathbf{T}_i \in \Theta$ for all $i \leq k$. As long as your method is consistent, it can be forced to return the various theories $\mathbf{T}_1, \mathbf{T}_2$, and so on, with arbitrarily high probability as sample size increases. It does not matter how clever or subtle your consistent method is at "mining" empirical samples—in fact more sensitive and informative data-mining techniques will "leap" for the successive theories in the empirical approximation chain more quickly than dull, insensitive ones. Moreover, it turns out that there is no way to bound the time at which these theories are produced—they happen when nature chooses, if the method is paying attention. These ideas are made precise in the following proposition, which is the central lemma for the causal flipping theorem:

**Proposition 2.** *Suppose that $M$ is consistent for $\Theta$ and that there exists chain*

$$\mathbf{T}_0 \leq \mathbf{T}_1 \leq \ldots \leq \mathbf{T}_k$$

*where $\mathbf{T}_i \in \Theta$ for all $i \leq k$. Let $p \in \mathbf{T}_0$ and let $\mathbf{O}$ be an arbitrary, open neighborhood of $p$ in total variation space. Let $\epsilon > 0$ and let $d$ be an arbitrary natural number. Then there exists $q \in \mathbf{O} \cap \mathbf{T}_k$ and there exist increasing sample sizes $n_0 < n_1 < \ldots < n_k$ such that for each $i \leq k$:*

$$n_i > d \cdot (i+1) \text{ and } q^{n_i}(M = \mathbf{T}_i) > 1 - \epsilon.$$

**Proof:** By induction on $k$. When $k = 0$, just choose $q$ to be $p$ and then $n_0 > d = d \cdot (0 + 1)$ exists by the consistency of $M$. At $k + 1$, there exists chain $\mathbf{T}_0 \leq \mathbf{T}_1 \leq \ldots \leq \mathbf{T}_{k+1}$. Let $p \in \mathbf{T}_0$ and let $\mathbf{O}$ be an open neighborhood of $p$ in total variation space. Since $M$ is consistent, there exists $n_0 > d$ such that:

$$p^{n_0}(M = \mathbf{T}_0) > 1 - \epsilon.$$

There exists total variation ball $\mathbf{B}_p$ around $p$ such that for each $p'$ in $\mathbf{B}_p$, the preceding, strict inequality holds (again, $\rho$ bounds the probabilities of arbitrary events in the sigma field of the product measure induced by the probability measures under consideration). Hence, the inequality holds over all of $\mathbf{S} = \mathbf{O} \cap \mathbf{B}_p$, which is open. Since $\mathbf{T}_0 \leq \mathbf{T}_1$, there exists $p' \in \mathbf{S} \cap \mathbf{T}_1$. By the induction hypothesis, there exists $q \in \mathbf{S} \cap \mathbf{T}_{k+1}$ and there exist increasing sample sizes $n_1, \ldots, n_{k+1}$ such that $n_i > 2 \cdot n_0 \cdot (i + 1)$ and:

$$q^{n_i}(M = \mathbf{T}_i) > 1 - \epsilon,$$

for each $i$ from 1 to $k + 1$. Thus, $n_i > d \cdot (i + 1)$, for each $i$ from 0 to $k + 1$. Since $q \in \mathbf{S} \subseteq \mathbf{B}_p$, we have that:

$$q^{n_0}(M = \mathbf{T}_0) > 1 - \epsilon.$$

Finally, $q \in \mathbf{S} \subseteq \mathbf{O}$. ⊣

If $A, B \in \text{FCIC}$, write $A \leq B$ if $\mathbf{F}_A \leq \mathbf{F}_B$. Say that $\mathbf{P}$ is **dependency driven** if and only if for all $A, B \in \text{FCIC}$,

$$B \subseteq A \Rightarrow A \leq B.$$

Dependency driven-ness is the second important concept in this paper. When $\mathbf{P}$ is dependency driven, statistical dependencies suffice to disambiguate any causal questions that can be disambiguated. As it happens, two of the most studied applications of causal discovery are dependency driven:

**Proposition 3.** *The set of linear Gaussian distributions over $\mathcal{V}$ and the set of discrete multinomial distributions over $\mathcal{V}$ are both dependency driven.*

The proof of the preceding proposition is presented in (Kelly and Mayo-Wilson 2010).[1] Recent work (Hoyer et al. 2009) suggests that non-dependency-driven questions may be more commonplace than we initially expected.

## 6 Retractions in Chance

In light of Proposition 2, any method $M$ probably can be forced, at successively large sample sizes, to produces $\mathbf{T}_1$, then produce $\mathbf{T}_2$, and so on, with arbitrarily high probability. These theories may be different. At each change of heart, the chance of producing the preceding theory drops severely (by at least the amount $1 - \epsilon$, where $\epsilon$ may be chosen as small as you please). We call that a **retraction in chance**. To quantify retractions in chance, let $\mathbf{P}$ be a set of measures on a common $\sigma$-field, $\Theta$ be a question, and let $M$ be a method, let $p \in \mathbf{P}$ and let $n > 0$. For two real numbers $x$ and $y$, define $x \ominus y = \max\{0, x - y\}$. Then define:

$$\begin{aligned}
r_\Theta(M, \mathbf{T}, p, n) &= p_{n-1}(M = \mathbf{T}) \\
&\quad \ominus p_n(M = \mathbf{T}); \\
r_\Theta(M, p, n) &= \sum_{\mathbf{T} \in \Theta} r_\Theta(M, \mathbf{T}, p, n); \\
r_\Theta(M, p) &= \sum_{n > 0} r_\Theta(M, p, n); \\
r_\Theta(M) &= \sup_{p \in \mathbf{P}} r_\Theta(M, p).
\end{aligned}$$

Then $r_\Theta(M, p)$ represents the total retractions in chance of $M$ in $p$ as sample size increases, and $r_\Theta(M)$ is the worst-case bound on $r_\Theta(M, p)$ with respect to $\mathbf{P}$. By tuning $\epsilon > 0$ in the preceding proposition arbitrarily low, one obtains the following lower bounds on retractions in chance:

**Proposition 4.** *Let the situation be as described in proposition 2, and suppose that $\mathbf{T}_i \neq \mathbf{T}_{i+1}$ for all $i < k$. Then $q$ satisfies $r_\Theta(M, q) > k - \epsilon$ and $r_\Theta(M) \geq k$.*

## 7 Causal Flipping Theorem

Suppose that $X \to Y$ is an edge in $\mathcal{G}$. Say that $X \to Y$ is **covered** if and only if the parents of $X$ in $\mathcal{G}$ are the same as those of $Y$ except for $X$ itself. A **covered edge reversal** involves flipping a covered edge to obtain a new $\mathcal{G}'$. Define $\mathcal{G} \preceq \mathcal{H}$ to hold if and only if $\mathcal{H}$ is obtained from $\mathcal{G}$ by a finite sequence of covered edge flips and edge additions. Then:

**Theorem 1** (Chickering 2002). *$I\mathcal{G} \subseteq I\mathcal{H}$ if and only if $\mathcal{H} \preceq \mathcal{G}$.*

Call a sequence $\mathcal{G}_0 \preceq \mathcal{G}_1 \preceq \ldots \preceq \mathcal{G}_k$ a **Chickering chain.** For example, the sequence in figure 1 is a Chickering chain. From (i) to (ii), start by adding edge $Z_1 \to Z_2$ and then flip the following edges in sequence: $Z_2 \to X$, $Z_1 \to X$, $X \to Y$, $Y \to Z_3$. Then add edge $Z_4 \to Y$. From (ii) to (iii), the process is similar, starting from the addition of $Z_3 \to Z_4$. We now present the main result of the paper. Say that a vertex $X$ is **isolated** in $\mathcal{G}$ if and only if $X$ has no causal connections to other variables in $\mathcal{G}$. Then:

**Theorem 2** (Causal Flipping). *Suppose that $M$ is consistent for the edge orientation question $\Theta_{X,Y}$ over dependency driven $\mathbf{P}$. Let $p \in \mathbf{P}$, $\mathcal{G} \in \mathbf{G}_p$, and suppose* there are at least k isolated variables in $\mathcal{G}$. Let **O** be an arbitrary, open neighborhood of $p$ in total variation space, and let $\epsilon > 0$. Then there exists $q \in \mathbf{O}$ such that $r_{\Theta_{X,Y}}(M, q) > k - \epsilon$. Thus, $r_{\Theta_{X,Y}}(M) \geq k$.

---

[1] The proof crucially employs the fact the set of unfaithful linear gaussian and discrete multinomial distributions each have Lebesgue measure zero; the linear gaussian case is discussed in (Spirtes, et. al 2000), and the discrete case in (Meek 1995).

**Proof:** It suffices to construct a Chickering chain $\mathcal{G} = \mathcal{G}_0 \preceq \mathcal{G}_1 \preceq \ldots \preceq \mathcal{G}_k$ in which causal edge $X - Y$ flips $k$ times. If $k = 0$, then the proposition is trivial, so suppose that $k > 0$. Suppose that DAG $\mathcal{G}$ contains the edge $X \to Y$, so that $\mathbf{T}_\mathcal{G}$ is either $\mathbf{T}_{X \to Y}$ or $\mathbf{T}_{X-Y}$. Extend $\mathcal{G}$ to some **complete** DAG $\mathcal{H}$ on the non-isolated vertices $\mathcal{V}'$ (i.e., a DAG in which each vertex is adjacent to every other). There exists a complete DAG $\mathcal{H}'$ over $\mathcal{V}'$ with edge $Y \to X$. Then $I\mathcal{H} = I\mathcal{H}'$, by proposition 1, and so $\mathcal{H}'$ results from $\mathcal{H}$ by a sequence of (covered) edge flips, by theorem 1. Let $Z$ be an isolated variable of $\mathcal{G}$. Add the edge $Z \to X$ to $\mathcal{H}'$ to obtain the DAG $\mathcal{G}_1$. Then, as $Z$ was isolated in $\mathcal{G}$, the vee $Z \to X \leftarrow Y$ is an unshielded collision in $\mathcal{G}_1$. By Proposition 1, the edge $Y \to X$ is essential in $\mathcal{G}_1$, and so $\mathbf{T}_{\mathcal{G}_1} = \mathbf{T}_{Y \to X}$. Hence $\mathbf{T}_{\mathcal{G}_1} \neq \mathbf{T}_\mathcal{G}$. Iterate this construction until the isolated variables are used up. ⊣

Then *every published causal conclusion $X \to Y$ drawn by a consistent method for the edge orientation question $\Theta_{X,Y}$ over dependency driven $\mathbf{P}_\mathcal{V}$ is subject to flipping any number of times, no matter how strong the evidence for that conclusion happens to be at present.* The bound $k$ on causal flips based on the number of isolated variables remaining in $\mathcal{V}$ is mainly a formal nicety. In practice, there are always new, unmeasured variables that have no noticeable correlation with the variables in $\mathcal{V}$, so the potential for future flips is, essentially, endless.

## 8 Causal Flips in Simulation

Suppose that the truth is the standardized structural equation model with independent, Gaussian error terms and causal path parameters fixed at the values depicted in Figure 2.

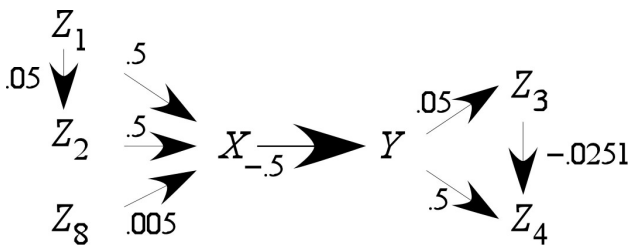

Figure 2: The Truth Behind the Simulations

We generated line graphs of frequency of output of the true edge $X \to Y$ (the pink "true" line) and the flipped edge $X \leftarrow Y$ (the blue "reversed" line) over 100 trials at each sample size examined, for each of four causal search algorithms implemented in TETRAD version 4.3.9-21.[2] The algorithms simulated are PC and FCI (Spirtes et al. 2000), CPC (Ramsey et al. 2006) and GES (Chickering et al. 2002). The retractions in chance of algorithms PC, CPC, and FCI in the causal scenario depicted in figure 2 are depicted in figures 3, 4, and 5. The GES algorithm is a bit hesitant to flip all the way to the false conclusion in the example depicted (the truth is still fully retracted), but on a slight refinement of the example in figure 2 (set $Z_3 \to Z_4$ to $-.02501$, $Z_8 \to X$ to $.005$, and $X \to Y$ to $.5$ - Reversing the sign on $X \to Y$ was an oversight that theoretically has nothing to do with the improved flips.) it bites enthusiastically for the first flip; the second flip being assured by the consistency of the GES method (figure 6).

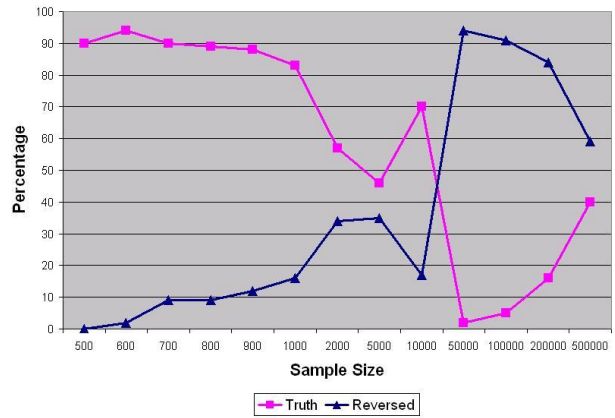

Figure 3: PC Algorithm

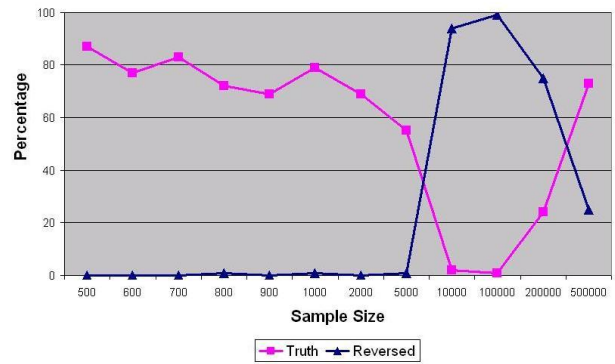

Figure 4: CPC Algorithm

---

[2]The TETRAD package is freely downloadable at www.phil.cmu.edu/projects/tetrad/ tetrad4.html.

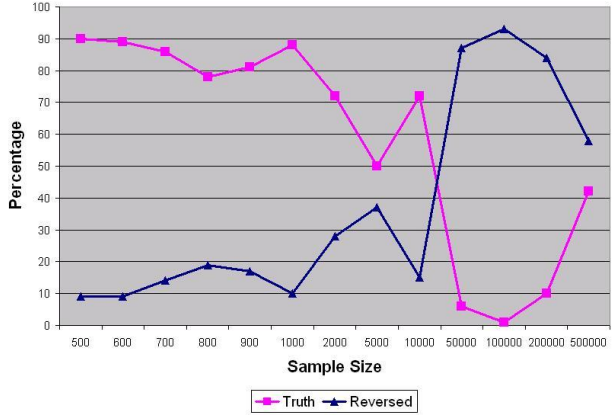

Figure 5: FCI Algorithm

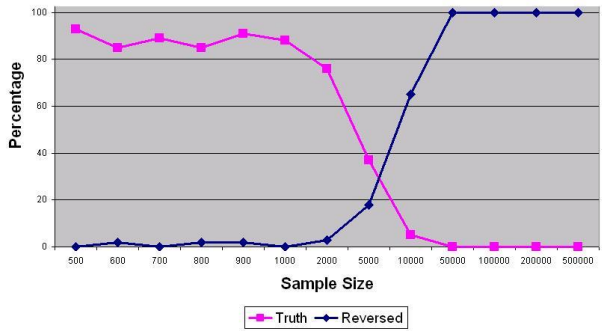

Figure 6: GES Algorithm

## 9 What Good are Methods that Flip?

Some advocates of causal discovery techniques invoke principles stronger than causal faithfulness that rule out the possibility of causal flips (Zhang and Spirtes 2003). Bayesians lay down prior probabilities that make the flips improbable. In contrast, we propose that the justification for existing causal discovery procedures is a *consequence* of the unavoidability of flipping, rather than assurances that flipping will not happen. Since repeated causal flips are unavoidable for consistent methods, the best that a truth-seeker can expect is to minimize such flips or retractions (Schulte et al. 2007). Hence, we propose that retraction minimization provides the best available explanation of how it is that causal discovery algorithms find true causes better than alternative strategies biased against simple causal networks can, at least in application to linear Gaussian and discrete causal networks. For comparisons of the retraction minimization approach with alternative foundations for model selection, see (Kelly 2010, Kelly and Mayo-Wilson 2008, Kelly 2008).

It can be demonstrated that, in both causal inference and some instances of scientific theorizing more generally, methods that minimize retractions are those that employ *Ockham's razor*, i.e. a systematic preference for simpler models (Kelly 2010, Kelly and Mayo-Wilson 2010, Kelly 2007b). The basic idea is that simpler theories imply fewer detectable **effects**, where effects are arbitrarily subtle phenomena that may me encountered arbitrarily late or only in arbitrarily large samples. Therefore, positing a theory that implies a potentially subtle effect that has not yet been detected leaves one open to having to retract the theory if the effect never appears (e.g., after a long sequence of failed or "null" attempts to produce the effects in question). That retraction gets added to all the unavoidable retractions that may follow, resulting in a sub-optimal retraction bound. So one ought not to produce complex theories that imply extra effects until those potentially subtle effects have been verified.

Some standard causal discovery procedures, like CPC and GES, do implement a version of Ockham's razor because they employ a systematic bias toward causal networks with fewer edges. That explains the nearly optimal performance of CPC and GES in terms of retractions (i.e. two full retractions in chance) in the above simulations. In contrast to CPC and GES, the PC and FCI algorithms violate Ockham's razor, which is why they exhibit "spikes" in the chance of producing orientation $X \to Y$ around sample size 10,000. The PC algorithm, for example, seeks to minimize conditional independence tests by deducing some CICs from others, assuming that the data are faithful - which can easily be false due to sample variation. As a consequence, PC's computational heuristics result in less direct convergence to the truth, where "directness" is measured in terms of total retractions in chance.

Here's it how happens. At sample size 10,000, the PC algorithm checks the independence of $X$ and $Z_3$ conditional on $\{Z_1\}$, $\{Z_2\}$, and $\{Z_8\}$. The tests fail to reject about $1/3$ of the time. In those cases, the algorithm *deduces*, from the false assumption that the data are faithful, that there is a collision $Z_3 \to Y \leftarrow X$.[3] The algorithm concludes, therefore, that the truth is $X \to Y$ when $X \leftarrow Y$ is compatible with the data *and* is equally simple. Ockham's razor (and common sense) demand that science wait for the *data* to choose among equally simple models.

The penalty for violating Ockham's razor is *not* unreliability—reliability is impossible in this question, in any event, unless one adopts unrealistic background assumptions stronger than faithfulness. However, there *is* a demonstrable penalty for violating Ock-ham's razor: the additional retraction in chance incurred on the downside of the probability "spike". In contrast, the CPC algorithm does cross-check in the recommended manner, and as a result, it avoids violating Ockham's razor in the manner just described. Thus, it avoids the spike, as does the score-based GES method. Therefore, the simulation study concretely illustrates, with real methods, how the retraction minimization theory can motivate and justify concrete improvements in data mining technology.

---

[3]We are indebted to Joe Ramsey for this explanation.

**Acknowledgements**

This material is based upon work supported by the National Science Foundation (NSF) under DUE Grant No 0740681. Any opinions, findings and conclusions or recommendations expressed in this material are those of the author(s) and do not necessarily reflect the views of the NSF.

This paper benefited from detailed criticisms by Clark Glymour. The authors are also indebted to Peter Spirtes, Joe Ramsey, Richard Scheines, for theoretical discussions and help with TETRAD. The TETRAD interface, implemented by Joe Ramsey, is extremely well-designed and facilitated our simulation studies. We would also like to thank Cosma Shalizi for comments concerning a related paper at the Formal Epistemology Workshop FEW.

**References**

Chickering, D. (2002) "Optimal Structure Identification With Greedy Search," *Journal of Machine Learning Research* 3: pp. 507-554.

Hoyer, P. O., D. Janzing, J. Mooij, J. Peters, and B. Schlkopf. (2009) "Nonlinear causal discovery with additive noise models." *Advances in Neural Information Processing Systems* 21 (NIPS 2008), pp. 689-696.

Kelly, K. (2010) "Simplicity, Truth, and Probability", forthcoming, *Handbook on the Philosophy of Statistics,* Malcolm Forster and Prasanta Bandyopadhyay, eds. Dordrecht: Kluwer.

Kelly, K. (2008) "Ockham's Razor, Truth, and Information," in *Philosophy of Information,* Van Benthem, J. Adriaans, P. eds. Dordrecht: Elsevier, pp. 321-360.

Kelly, K. (2007a) "Ockham's Razor, Empirical Complexity, and Truth-finding Efficiency," *Theoretical Computer Science.* 383: 270-289.

Kelly, K. (2007b) "A New Solution to the Puzzle of Simplicity," *Philosophy of Science* 74: 561-573.

Kelly, K. and C. Mayo-Wilson. (2008). Review of *Reliable Reasoning* by Gilbert Harman and Sanjeev Kulkarni. *Notre Dame Philosophical Reviews.* Available electronically at http://ndpr.nd.edu/review.cfm?id=12684

Kelly, K. and C. Mayo-Wilson. (2010) "Ockham Efficiency Theorem for Random Empirical Methods," forthcoming, *Journal of Philosophical Logic.*

Kelly, K. and C. Mayo-Wilson. (2010) "The Causal Flipping Theorem" Technical Report 189, Department of Philosophy, Carnegie Mellon University.

Meek, C. (1995). "Strong completeness and faithfulness in Bayesian networks." *Proceedings of the Eleventh Annual Conference on Uncertainty in Artificial Intelligence*, pp. 411-418

Pearl, J. and Verma, T (1991), "A Theory of Inferred Causation," in J.A Allen, R. Fikes, and E. Sandewall eds., *Principles of Knowledge Representation and Reasoning: Proceeding of the Second International Conference*, San Mateo, CA: Morgan Kaufmann, pp. 441-452.

Ramsey, J., Zhang, J., and Spirtes, P. (2006), "Adjacency-Faithfulness and Conservative Causal Inference", *Proceedings of the 22nd Annual Conference on Uncertainty in Artificial Intelligence*, pp. 401-408.

Richardson, T. and Spirtes, R. (2002) "Ancestral Graph Markov Models," *Annals of Statistics* 30: pp. 962-1030.

Robins, J., Scheines, R., Spirtes, P., and Wasserman, L. (1999) "Uniform Consistency in Causal Inference," *Biometrika* 90:491-515.

Schulte, O., Luo, W., and Griner, R. (2007) "Mind Change Optimal Learning of Bayes Net Structure". In *20th Annual Conference on Learning Theory* (COLT), San Diego, CA, June 12-15.

Shimizu, S., P. O. Hoyer, A. Hyvarinen, and A. J. Kerminen. (2006) "A Linear Non-Gaussian Acyclic Model for Causal Discovery." *Journal of Machine Learning Research.* 7: pp. 20032030.

Spirtes, P., Glymour, C., and Scheines, R. (2000) *Causation, Prediction, and Search.* Cambridge: M.I.T. Press.

Zhang, J. and Spirtes, P. (2003) Strong Faithfulness and Uniform Consistency in Causal Inference, in *Proceedings of the Nineteenth Conference on Uncertainty in Artificial Intelligence*, 632-639. Morgan Kaufmann.